\documentclass[letterpaper, 10 pt, conference]{ieeeconf}
\usepackage[binary-units=true]{siunitx}
\usepackage{times}
\usepackage{amsmath,amssymb}
\usepackage[nolist,nohyperlinks]{acronym}
\usepackage{accents}
\usepackage{graphicx}
\usepackage{bm}
\usepackage{bbm}
\let\labelindent\relax
\usepackage[inline]{enumitem}
\usepackage{array}
\usepackage{multicol}
\usepackage{multirow}
\usepackage{tabulary}
\usepackage{booktabs}
\usepackage{subfig}
\usepackage{todonotes}
\usepackage{cite}
\usepackage{algorithm}
\usepackage{algpseudocode}
\usepackage[bookmarks=true]{hyperref}

\newcommand{\norm}[1]{\left\lVert#1\right\rVert}
\newtheorem{remark}{Remark}

\begin{acronym}
	\acro{CoM}{Center of Mass}
	\acro{RL}{reinforcment learning}
	\acro{IL}{imitation learning}
	\acro{MIP}{Mixed-integer Programming}
	\acro{MIQP}{Mixed-integer Quadratic Program}
	\acro{RNN}{Recurrent Neural Network}
	\acro{CNN}{Convolutional Neural Network}
	\acro{DDP}{Differential Dynamic Programming}
	\acro{MC}{Motion Capture}
	\acro{MLP}{Multilayer Perceptron}
	\acro{MCTS}{Monte Carlo Tree Search}
	\acro{MDP}{Markov-decision Process}
	\acrodefplural{MDP}{Markov-decision Processes}
	\acro{MPC}{model predictive control}
	\acro{ADMM}{Alternating Direction Method of Multipliers}
	\acro{BC}{behavioral cloning}
	\acro{NN}{neural network}
	\acro{QP}{Quadratic Program}
	\acrodefplural{QP}{Quadratic Programs}
	\acro{DoF}{Degree of Freedom}
	\acro{VOCAM}{Vision-Based Online Cost Adaptation for Model Predictive Control}
	\acro{OCP}{Optimal Control Problem}
	\acro{IOC}{inverse optimal control}
	\acrodefplural{DoF}{Degrees of Freedom}%
			
\end{acronym}
\IEEEoverridecommandlockouts          
\renewcommand{\baselinestretch}{0.9786}

\title{\LARGE \bf MPC with Sensor-Based Online Cost Adaptation}

\author{Avadesh Meduri$^{1}$, Huaijiang Zhu$^{1}$, Armand Jordana$^{1}$, Ludovic Righetti$^{1}$%
\thanks{$^{1}$Tandon School of Engineering, New York University, USA}%
\thanks{This work was in part supported by the National Science Foundation (grants 1825993, 1925079, 2026479 and 1932187) and Meta Platforms Inc.}% <-this % stops a space
}

\begin{document}
\maketitle
\thispagestyle{empty}
\pagestyle{empty}

%===============================================================================

\begin{abstract}
Model predictive control is a powerful tool to generate complex motions for robots. However, it often requires solving non-convex problems online to produce rich behaviors, which is computationally expensive and not always practical in real time. Additionally, direct integration of high dimensional sensor data (e.g. RGB-D images) in the feedback loop is challenging with current state-space methods.
This paper aims to address both issues. It introduces a model predictive control scheme, where a neural network constantly updates the cost function of a quadratic program based on sensory inputs, aiming to minimize a general non-convex task loss without solving a non-convex problem online. By updating the cost, the robot is able to adapt to changes in the environment directly from sensor measurement without requiring a new cost design. Furthermore, since the quadratic program can be solved efficiently with hard constraints, a safe deployment on the robot is ensured. Experiments with a wide variety of reaching tasks on an industrial robot manipulator demonstrate that our method can efficiently solve complex non-convex problems with high-dimensional visual sensory inputs, while still being robust to external disturbances.
\end{abstract}

%===============================================================================
\section{Introduction}
Robots deployed in factories and warehouses are able to perform repetitive tasks accurately. However, they are not able to carry out more complex tasks as they lack the ability to adapt their motions according to high-dimensional sensory feedback such as vision, touch, or sound from the surrounding environment. This inadequacy arises from the fact that these robots typically track pre-defined motions that do not adapt to changes in the environment.

%  The behaviours needed to achieve these tasks on robots are usually generated using trajectory optimization techniques [citation]. These methods can generate various motions with limited human input and have improved robot autonomy.

%% GOAL - To explain limitations and advantages of numerical methods 
%% Not clear how to handle high dimensional data

One popular way to adapt motions online is through \ac{MPC}, which requires solving trajectory optimization problems repeatedly based on state feedback. MPC has become standard in many robotic applications, such as legged locomotion, enabling robots to adapt their behavior to unforeseen events. 
Many approaches use linear approximations of the non-linear robot dynamics and quadratic cost to describe the desired behaviours such as walking, trotting or bounding ~\cite{herdt2010online, kim2019highly, khadiv2020walking}. This then allows the trajectory optimization problem to be solved as a \ac{QP}. Thus, they guarantee real-time convergence to a unique global optimum and hard constraint satisfaction, leading to robustness to external disturbances and easy transfer to robots. However, they can only generate limited types of behaviours due to the use of simplified models and fixed, hand-tuned cost functions that do not change based on sensor feedback. As a result, different behaviours often require different hand-crafted cost functions and simplified models.  

Recent advances in \ac{MPC} have removed the need for reduced order models by directly using the full non-linear robot dynamics~\cite{kleff2021high, koenemann2015whole, meduri2022biconmp, mastalli2022agile, erez2013integrated, neunert2018whole}. This enables rapid generation of a variety of behaviours with a unified formulation. However, low computation time is achieved by making sacrifices such as inability to enforce hard constraints~\cite{kleff2021high, neunert2018whole, mastalli2022agile} or allowing only quadratic costs~\cite{meduri2022biconmp}. Consequently, any increase in the complexity of the cost, such as adding non-convex obstacle avoidance tasks, would significantly increase computation time. To date, only empirical real-time convergence has been shown with these methods. 

More importantly, visual sensory data is rarely used in the feedback loop, because modelling system or environment dynamics with images as part of the state is usually intractable. 
% Even if such visual dynamics were obtained through learning-based methods~\cite{watter2015embed}, it would not be straightforward to incorporate them in a real-time MPC loop because they are highly nonlinear and computing their derivatives through auto-diff is too slow for real-time use cases. 
As a result, the few approaches that do use visual inputs extract important features from the images such as obstacles, goal locations or keypoints through image processing~\cite{pankert2020perceptive} or train neural networks that checks for collisions with the environment~\cite{bhardwaj2022storm}. This information is then included manually inside the cost function. Consequently, the feature extraction is tailored for a particular task and would require modifications for new situations. Further, image processing based methods are sensitive to lighting, image contrast, etc. which can undermine the quality of trajectories or stability of the numerical solver, especially when the underlying optimizer cannot enforce hard constraints \cite{pankert2020perceptive}.

To address these issues, we introduce a \ac{MPC} scheme featuring an adaptive \ac{QP}. The cost function of this \ac{QP} is constantly updated by a neural network directly from multiple sensors (e.g. joint encoders and vision) and aims to generate trajectories that minimize a non-convex general task loss. This procedure enables solving a convex problem (\ac{QP}) online while still generating rich behaviours defined by a non-linear general task loss. Consequently, this guarantees safe deployment of sensor-driven \ac{MPC} on real robots by enforcing hard constraints and real-time convergence. At the same time, direct integration of vision into the \ac{MPC} feedback loop is simple by leveraging the neural network. Thus eliminating the need for hand-engineered feature extraction.

\begin{figure*}
%\vspace{0.2cm}
\centering
\includegraphics[scale=0.55]{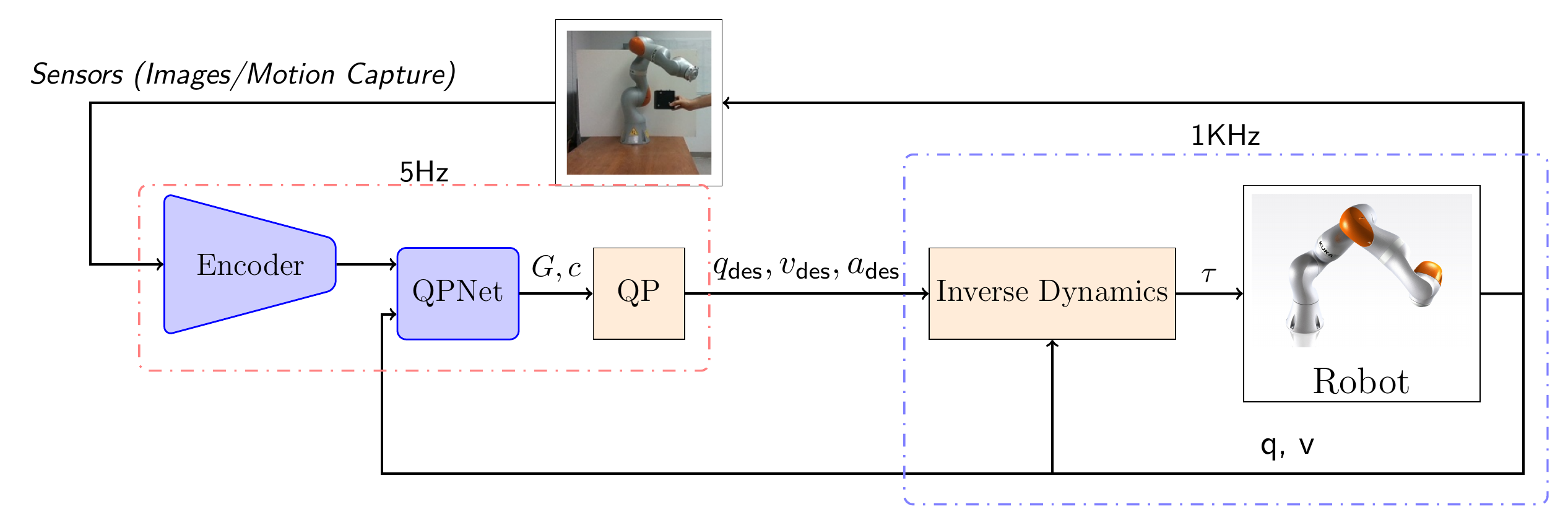}
\caption{Pipeline in the case of image measurement. Note that in the case of measurement from motion capture, the goal position is directly provided to the QPNet (encoder is an identity map).}
\label{fig:pipeline}
\vspace{-0.5cm}
\end{figure*}

Figure~\ref{fig:pipeline} depicts the MPC scheme deployed on the robot. For every new sensor measurement, a neural network, called QPNet, predicts the cost function. Then, the associated \ac{QP} is solved and its solution is tracked by a model-based inverse dynamics controller at a higher frequency. Consequently, at each new sensor measurement, the \ac{QP} varies and adapts to changes in the environment and the desired task goal, in order to minimize the general non-linear task loss. The online adaption of the cost function allows the generation of complex behaviours while still solving a \ac{QP}. The training of the QPNet is done in a supervised way after generating the \ac{QP} costs using techniques of implicit differentiation~\cite{amos2017optnet}.

The proposed method is a general reactive \ac{MPC} framework that has three main advantages: 
\begin{enumerate}
    \item Allows a simple and clean way to incorporate high-dimensional sensory inputs into the feedback loop while still using established optimization techniques for \ac{MPC}.
    \item Enforces hard constraints during deployment even while using neural networks in the feedback loop to integrate different sensor modalities. 
    \item Guaranteed real-time solve rates of the MPC loop that remain constant irrespective of the task complexity. 
\end{enumerate}
We demonstrate these advantages by choosing one application - a set of reaching tasks (3D locations and 6D poses) on a $7$-DoF KUKA LBR iiwa robot under different settings 
\begin{enumerate*}
    \item when the target location is provided by a \ac{MC} system,
    \item when only a RGB-D camera is available, and
    \item when the robot is required to avoid an obstacle.
\end{enumerate*}
We also demonstrate stable push recovery in all the above mentioned situations. 
These experiments show that by simply modifying the high-level task loss, our approach can handle more complex settings without any change to the general method while retaining all the above mentioned advantages. This is in contrast to existing methods for adaptive reaching tasks that only show a subset of these advantages and require modification in the cost design to extract suitable features from images~\cite{pankert2020perceptive, bhardwaj2022storm}.  
% For e.g, \ac{DDP} based methods can not enforce constraints and need image processing . Fast sampling based methods do not have convergence guarantees, use hand designed integration of learned cost for obstacle avoidance and do not show robustness to pushes \cite{bhardwaj2022storm}. 
%
% To our knowledge, this is the first demonstration in the literature of stable, robust and high frequency MPC on a real robot by using RGB-D images in the feedback loop with guaranteed safe deployment (enforces hard constraints) and real-time convergence regardless of task complexity.

% \begin{remark}
% The Bi-level formulation enables capturing the complex robot dynamics in the cost function while still using double integrator dynamics. 
% \end{remark}

\section{Related Work}
\paragraph{Learning-based \ac{MPC}}It is not a new idea to approximate \ac{MPC} with neural networks. For example,  Parisini and Zoppoli have proposed in~\cite{parisini1995receding} to approximate a receding-horizon controller (\ac{MPC}) for a nonlinear system using neural networks almost three decades ago. Since then, advances have been made by exploiting techniques from explicit \ac{MPC} to establish constraint satisfaction~\cite{chen2018approximating} and robust \ac{MPC} to improve robustness by design~\cite{nubert2020safe}. However, they do not address the problem of fusing high-dimensional sensory data. More recently, this methodology has been further investigated through the lens of \ac{IL}. In particular, the \ac{MPC} policy is viewed as an algorithmic expert to be imitated by a student policy that can incorporate high-dimensional sensory inputs~\cite{pan2017agile}; however, in this work, the underlying trajectory optimization problem is solved by dynamic differential programming (DDP), thus does not enforce hard constraints.

\paragraph{Implicit differentiation through convex optimization} A key ingredient of our approach is the ability to differentiate through convex optimization problems via the implicit function theorem~\cite{amos2017optnet, agrawal2019differentiable}. Indeed, such techniques have been successfully applied to learning parameters of optimization-based control policies. For instance, Amos et al. learns the cost function and the dynamics jointly for a \ac{MPC} policy in a model-free setting~\cite{amos2018differentiable}; Agrawal et al. tunes parameters for various forms of control policies represented as convex optimization~\cite{agrawal2020learning}. However, neither of these methods address the problem of fusing high-dimensional sensory data and they have only been demonstrated on simple problems without deploying the learned policies on real hardware.

\paragraph{Inverse optimal control} Learning cost functions to generate desired behaviours is studied in \ac{IOC}. In particular, cost parameters are inferred to recreate desired behaviour from expert demonstrations with policy optimization~\cite{kalakrishnan2013learning, ratliff2006maximum, abbeel2004apprenticeship}. Visual demonstrations have also been used to learn cost functions via bi-level optimization~\cite{das2020model}. However, in this line of work, the learned cost cannot be adapted by additional sensory inputs to generalize to new situations. Also, due to the use of trajectory optimization, our approach does not require human demonstrations, which can be tedious to obtain.

\section{Method} \label{sec:Method}
In this section, we present our approach that is able to perform \ac{MPC} to achieve a non-convex reaching task~\cite{kleff2021high} with high-dimensional sensory inputs in real time on a robot. 

\subsection{Problem Formulation} \label{sec:problemformulation}

We consider a typical optimal control problem, where the goal is to find an optimal trajectory $x^{\star}$ of states and controls that minimizes a non-convex task loss. Instead of directly solving this problem with a nonlinear optimization method, we would like to re-parameterize it as a \ac{QP} so that it can be solved efficiently at run time. To achieve this, we formulate the trajectory optimization problem as bi-level optimization, with an upper-level non-convex cost that describes the complex motion needed to be performed and a lower-level \ac{QP}-based trajectory optimizer. Mathematically, the problem is formulated as follows
% We consider the robot state to be joint positions and velocities $s_t=[q_t, v_t]$ and the control input is the joint acceleration $u_t = a_t$. Given an initial state $s_{\text{init}} = [q_{\text{init}}, v_{\text{init}}]$, we would like to find a trajectory $x^{\star} = [q_0^{\star}, v_0^{\star}, a_0^{\star},\dots,q_N^{\star},v_N^{\star}]$ to realise a task that is described by a higher level non-convex cost function. However, 
\begin{align}
& \underset{G, c}{\min} \quad \quad \quad  \phi_{\text{task}}(x^{\star}, g) \label{eq:task_loss}\,,
\end{align}
\vspace{-0.2cm}
where
\begin{subequations}\label{eq:genericQP}
\begin{align} 
x^{\star} = \quad &\underset{x}{\operatorname{argmin}}  \quad \quad x^{T} G x + c^{T} x \label{eq:lower_level_loss}\\
& \text{s.t.} \quad \quad  Ax = b,\,\, Kx \leq h\,. \label{eq:constraints}
\end{align}
\end{subequations}
This bi-level optimization problem aims to minimize the high-level non-convex task loss~\eqref{eq:task_loss} parameterized by a task goal $g$ with respect to the cost parameters, the matrix $G$ and the vector $c$, of the lower-level \ac{QP}~\eqref{eq:genericQP}. The \ac{QP} has a formulation of a standard optimal control problem, where the constraints~\eqref{eq:constraints} enforce the system dynamics and the feasibility of the states and controls.  

We solve this bi-level problem with Adam~\cite{kingma2014adam}. The gradients through the lower-level \ac{QP} ($\frac{\partial x^{\star}}{\partial G }$ and $\frac{\partial x^{\star}}{\partial c }$), are obtained using the implicit differentiation technique proposed in \cite{amos2017optnet}. The cost function matrix $G$ is internally parameterized as $G = \textit{ReLu}(L) \textit{ReLu}(L)^{T}$ where $L$ is the actual parameter optimized in the bi-level problem. This ensures that all constructed cost parameters $G$ in the \ac{QP} are positive semidefinite. Since the dynamics is known and the gradients are exact, our method does not require sampling to estimate the gradient as opposed to model-free methods~\cite{das2020model, levine2016end} and takes very few iterations to converge.  

\begin{remark}
It is important to note that we design the lower-level problem as a \ac{QP} due to the availability of efficient off-the-shelf solver, bounded computation time for reliable deployment, and guaranteed convergence to global optimum (which is important for implicit differentiation). However, the lower-level problem can also be represented as cone programming or geometric programming~\cite{agrawal2019differentiable}, which are differentiable as well. 
\end{remark}

\subsection{Trajectory Optimization for a Reaching Task}
We now use the above optimization problem to perform various reaching tasks with a robot manipulator. In this scenario, the goal $g$ is a $3$D vector describing the target reaching position of the end-effector. For this, we would like to generate a trajectory $x$ of joint positions $q$, velocities $v$ and accelerations $a$ that minimize the high-level non-convex task loss~\eqref{eq:task_loss}. Consequently, we set the \ac{QP}-based trajectory generator to
\begin{subequations}
\label{eq:qp_invdyn}
\begin{align}
\operatornamewithlimits{min}_{x=[q_{0}, v_{0}, a_{0}, \dots q_{N}, v_{N}]} x^{T} G x + c^{T} x \label{eq:IK} \\
\text{s.t. } v_{t+1} = v_{t} + a_{t}\delta t , 
\quad q_{t+1} = q_{t} + v_{t+1}\delta t\label{eq:double_integrator} \\
a_{\text{min}} \leq a_{t} \leq a_{\text{max}}, 
\quad q_{0}, v_{0}  = q_{\text{init}}, v_{\text{init}}\label{eq:limits}
\end{align}
\end{subequations}
% \begin{align}
%   \underset{x=[q_{0}, v_{0}, a_{0}, \dots q_{N}, v_{N}]}{\operatorname{min}}  &\quad \quad x^{T} G x + c^{T} x \label{eq:IK}\\
%  \text{s.t.} \quad \quad & v_{t+1} = v_{t} + a_{t}\delta t , 
% \quad q_{t+1} = q_{t} + v_{t+1}\delta t\label{eq:double_integrator} \\  
% & a_{\text{min}} \leq a_{t} \leq a_{\text{max}}, 
% \quad q_{0}, v_{0}  = q_{\text{init}}, v_{\text{init}}\label{eq:limits}
% \end{align}
where $a_{\text{min}}, a_{\text{max}}$ are the acceleration limits (which indirectly enforce torque limits) and $q_{\text{init}}, v_{\text{init}}$ is the initial state for the \ac{MPC} optimization problem. This \ac{QP} generates a feasible joint trajectory $x^{\star} = [q^{\star}_{0}, v^{\star}_{0}, a^{\star}_{0},\dots,q^{\star}_{N},v^{\star}_{N}]$ that minimizes the quadratic cost defined by $G,c$. % Consequently, we use the bi-level optimization method discussed previously to compute the optimal cost functions to reach different goal positions for a given $q_{\text{init}}, v_{\text{init}}$.

\begin{remark}
Note that while the \ac{QP} only enforces double integrator dynamics, the complex robot model is implicitly incorporated in the convex costs $G,c$. This is because the upper-level cost in the bi-level optimization contains the non-linear robot kinematics. Additionally, an inverse dynamics controller takes into account the robot dynamics to realize the desired joint commands (cf. Sec.~\ref{sec:execution}).
\end{remark}

\subsection{Online Cost Adaptation using Measurement Data} \label{sec:training}

With the bi-level optimization method discussed in Sec.~\ref{sec:problemformulation}, it is now possible to find cost parameters $y^{\star} = [G^{\star}, c^{\star}]$ that minimize the non-convex cost~\eqref{eq:task_loss} via \ac{QP} given an initial state $s_{\text{init}} = [q_{\text{init}}, v_{\text{init}}]$  and a goal position $g$. While the \ac{QP} can be solved efficiently, the bi-level problem~\eqref{eq:task_loss} is still non-convex; hence, it does not have any real-time guarantees and might not be solved fast enough if the task loss is very complex. Furthermore, at run time, the goal position $g$ might not be directly available and one might only have access to sensor data describing $g$. Therefore, we learn a policy network $\hat{\pi}$ which we refer to as QPNet, mapping $s_{\text{init}}$ and sensor information to the optimal \ac{QP} cost parameters $G^{\star}, c^{\star}$. This ensures that at run time, given sensor data, the cost parameters are available in a bounded amount of time from a neural network forward pass.
In this work, we use two different types of measurements of the goal (reaching position): \ac{MC} and RGB-D images.

\paragraph{\ac{MC} measurement} In this setting,  the QPNet takes as an input the \ac{MC} measurement. It is assumed that the \ac{MC} system directly provides the ground truth position of the desired reaching location $g$ (the center of the cube in Fig.~\ref{fig:pipeline}). This means that to train the QPNet, no real sensor data is needed.  To create the dataset, an appropriate task loss is chosen based on the desired task (e.g. reaching, reaching with obstacle etc.). We then randomly sample initial robot states~$s_{\text{init}}$ and desired reaching positions~$g$. Then, the bi-level problem~\eqref{eq:task_loss} is solved in a \ac{MPC} fashion until the goal is reached. That is, at each \ac{MPC} step, the last state $q,v$ of the computed trajectory from the \ac{QP} is used as the initial state for the next iteration. All the pairs $\{(s_{\text{init}}, g), y^{\star}\}$ collected along this simulated trajectory are then added to the dataset. This gives us a dataset containing approximately $10, 000$ samples that is used to train the QPNet in a supervised manner with a $l_1$-loss: $ \norm{y^{\star} - \hat{\pi} (s_{\text{init}}, g)}_{1} $. This can be seen as \ac{BC}, a naive form of \ac{IL}. However, the choice of the specific \ac{IL} algorithm is not restricted by our method; it is entirely possible to use direct policy learning approaches such as DAgger~\cite{ross2011reduction} since we can always query the trajectory optimizer for more data.

% using a \ac{MC} system is often infeasible when deploying robots in real environments and it is desirable to work directly with images/cameras during execution
\paragraph{Image measurement}
Ultimately, it is expensive to use a \ac{MC} system for each of the many robots during deployment. Instead it is preferable to use the \ac{MC} system once, to train a vision system and then use the trained system on several cameras for several robots. Integrating vision can be achieved in our method by encoding RGB-D images to a lower-dimensional embedding~$e$, which is then given as an input to the QPNet. We present here a procedure to train the encoder alongside the QPNet using \ac{MC} measurement. To obtain a suitable encoder for the reaching task, we first train a \ac{CNN} to predict the location of the cube given an RGB-D image. The \ac{CNN} is trained on $150,000$ pairs of RGB-D image and ground truth cube location obtained from a \ac{MC} system. After training, the convolution layers are frozen and used as an encoder that outputs the encoding $e$ given an image of the cube. 
To train the QPNet that can directly take the encoding $e$ as an input, we use the previous dataset and generate $10,000$ pairs of the form  $\{ (s_{\text{init}}, e), y^{\star} \}$. Here, $e$ is obtained by inputting the images from the previous dataset into the trained encoder; $y^{\star} = [G^{\star}, c^{\star}]$ is obtained by solving problem~\eqref{eq:task_loss} using the goal $g$ that is already obtained by \ac{MC} when collecting the image dataset; and lastly, $s_{\text{init}}$ is sampled randomly. This results in a vision-based policy $\hat{\pi}_{\text{vision}}(s_{\text{init}}, e)$ that can fuse the RGB-D sensor data into the feedback loop, whereas traditional trajectory optimization methods would require processing the image into a amenable form~\cite{pankert2020perceptive, bhardwaj2022storm}. 
% (e.g. by adding constraints based on extracted image features). Further, as discussed in Sec~\ref{sec:problemformulation}, hard constraints in the original problem can still be enforced even when RGB-D images are used in the feedback loop. 

\begin{remark}
We train the encoder from scratch because collecting $150$k image at $60$ FPS took us $\approx 40$ minutes, while training the CNN took an hour (the QPNet still requires only $10$k samples). To further reduce the encoder training time and improve robustness, a pre-trained encoder~\cite{he2022masked, nair2022r3m} could be used and fine-tuned on our data. Another option is to train the encoder in simulation where privilege information is readily available. In this case, the sim2real transfer can be considered as a separate problem where promising progress is being made. We have not explored these directions because our main focus in this work is to propose a stable \ac{MPC} algorithm that can include vision feedback. 
\end{remark}

\subsection{Multi-Modal Sensor-Based MPC Loop on the robot} \label{sec:execution}
At run time, the sensor measurement (either the \ac{MC} measurement or the image encoding) and the robot state $q,v$ are provided to the QPNet which predicts the \ac{QP} costs. With this prediction, a \ac{QP} is then constructed and solved in order to obtain the desired joint trajectory ${x_{\text{des}} = [q^{0}_{\text{des}}, v^{0}_{\text{des}}, a^{0}_{\text{des}},\dots,q^{N}_{\text{des}},v^{N}_{\text{des}}]}$, a segment of which is interpolated and tracked by an inverse dynamics controller. The torques $\tau$ needed to realise this trajectory are computed using ${\tau = RNEA(q,v,a_{\text{des}}) + k_{p}(q - q_{\text{des}}) + k_{d}(v-v_{\text{des}})}$, where $RNEA$ is the Recursive Newton Euler Algorithm (inverse dynamics), $k_{p}, k_{d}$ are joint gains which are added for minor corrections and $q_{\text{des}}$,  $v_{\text{des}}$, $a_{\text{des}}$ are the desired joint position, velocity, and acceleration at the respective time step in the interpolated trajectory. Since the \ac{QP} enforces joint constraints, it is guaranteed that feasible joint torques can be computed for the given joint trajectory. Further, the use of a model-based inverse dynamics controller allows easy transfer to the robot and maintains compliance during deployment. 

The inverse dynamics controller runs at $\SI{1}{\kilo\hertz}$, while the sensory feedback loop runs at $\SI{5}{\hertz}$. The compute time of one pass through the QPNet and the QP is under $\SI{2}{\milli\second}$ and with the encoder it is under $\SI{10}{\milli\second}$, which means that if needed our approach could be run at $\SI{500}{\hertz}$ or $\SI{100}{\hertz}$. Figure~\ref{fig:pipeline} shows the entire control loop deployed on the robot in the case of using RGB-D image measurement.

\begin{remark}When using a fixed cost, the \ac{QP} can only generate limited behaviour. However, in our setup, by updating the cost function at each control cycle using direct sensor measurements, we are able to produce arbitrarily complex motions online.

% This is shown by using just a simple double integrator dynamics (that does not encode the robot model) and constantly adapting the cost function, complex reaching tasks can be achieved on the robot. Also, by incorporating more complexity in the task loss, such as using a differentiable simulator~\cite{werling2021fast, de2018end}, the framework can be used for applications beyond just reaching.

% While the requirement of using linear constraints in the lower-level problem may seem restrictive at first glance, it is actually possible to embed the non-linearity and non-convexity into the upper-level task loss. For example, via a differentiable simulator such as ~\cite{werling2021fast, de2018end}, we can unroll the lower-level solution and define more complex upper-level task loss that can reason about force interactions between the robot and the environment, which makes our framework applicable to well beyond simple reaching tasks.
\end{remark}

\section{Experimental Results} \label{sec:result}
In this section, we demonstrate the benefits of the proposed method on a $7$-DoF KUKA LBR iiwa robot for various reaching tasks. First, we show the performance of the QPNet that is trained to predict the \ac{QP} cost from low-dimensional sensor data: robot states $q,v$ and desired reaching position $g$ provided by \ac{MC}. Then, we show the ability of the approach to avoid obstacles while reaching different positions and 6D poses without any change to the computation time. Finally, we present results obtained when raw images of the cube (the desired reaching target) are used as an input to the pipeline along with the robot state $q, v$.

The neural networks are implemented using PyTorch~\cite{paszke2019pytorch}. Quadprog is used as the \ac{QP} solver for the lower-level problem. 
% The communication with the KUKA robot is achieved with a combination of the proprietary FRI software and our custom control stack. 
The RGB-D images are obtained using the Intel RealSense D435 camera at $60$ FPS and ground truth reaching locations are measured by a \ac{MC} system tracking the cube. 

\begin{figure}
%\vspace{0.2cm}
% \centering
\includegraphics[scale=0.11]{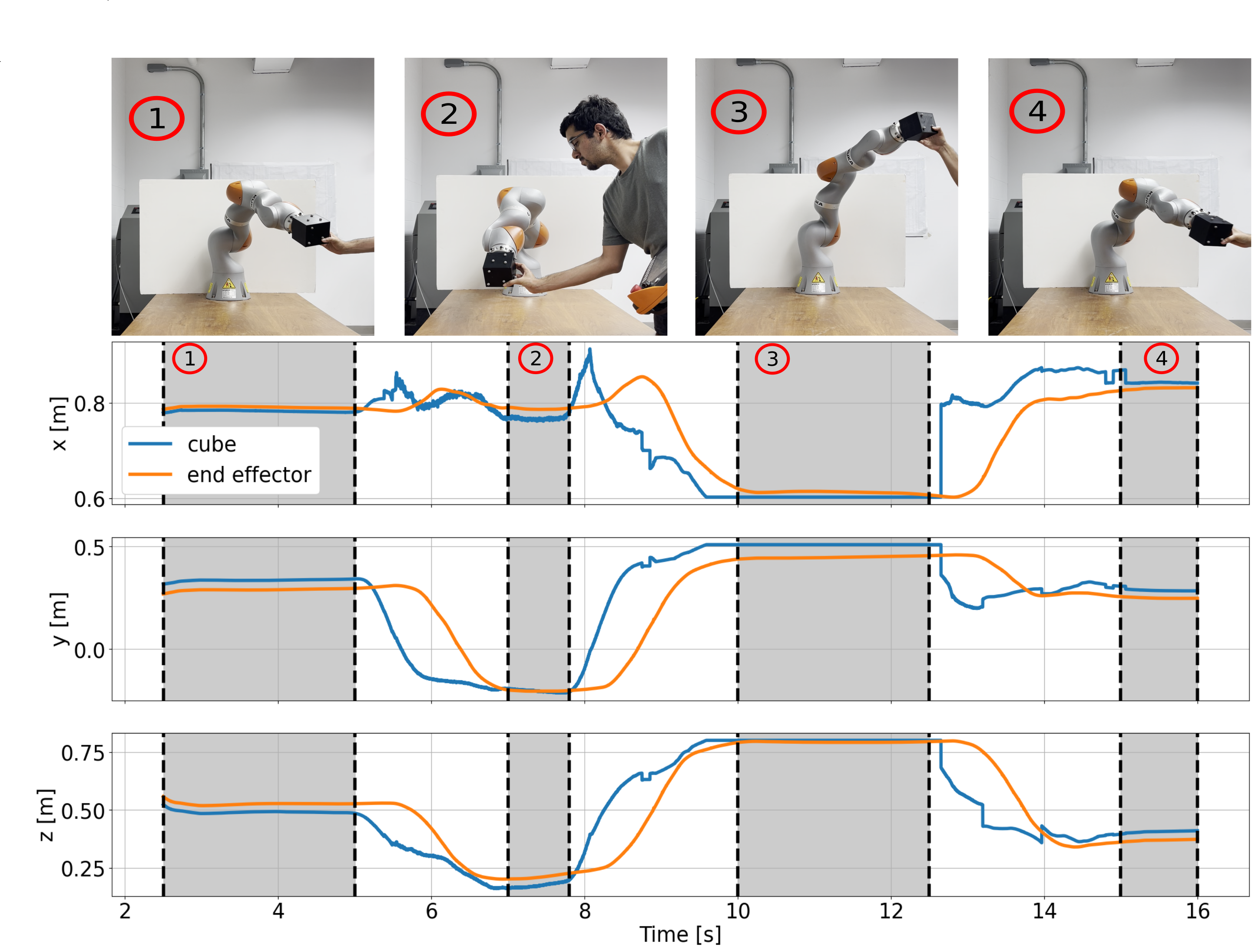}
\caption{KUKA reaching the cube at different locations. The numbers in the shaded area of the plot show the corresponding situation on the robot at that instant.}
\label{fig:vicon_reaching}
\vspace{-0.6cm}
\end{figure}

\subsection{Reaching Task with Motion Capture} \label{sec:vicon}

In this experiment, the goal for the KUKA robot is to reach a cube located in its workspace. The location of the cube is obtained using a \ac{MC} system. Therefore, we define a non-convex task loss $\phi_{\text{reaching}}(x,g)$ that aims to minimize the distance between the end-effector position and the desired goal position. In addition, the loss regularizes joint positions, velocities, accelerations and penalizes rapid changes in joint accelerations. The task loss is 
\begin{multline} \label{eq:reachingtask}   \phi_{\text{reaching}}(x, g) = w_{6}\norm{f_{\text{fk}}(q_{N}) - g} + w_{7}\norm{f_{\text{v}}(q_{N}, v_{n})} \\
            + \sum_{t=1}^{N-1} w_{1}\norm{f_{\text{fk}}(q_{t}) - g} + w_{2}\norm{q_{t}} + w_{3}\norm{v_{t}} +
            w_{4}\norm{a_{t}} \\ + w_{5}\norm{a_{t} - a_{t-1}}\
\end{multline}
\begin{figure*}[h]
\vspace{0.2cm}
\centering
\includegraphics[scale=0.32]{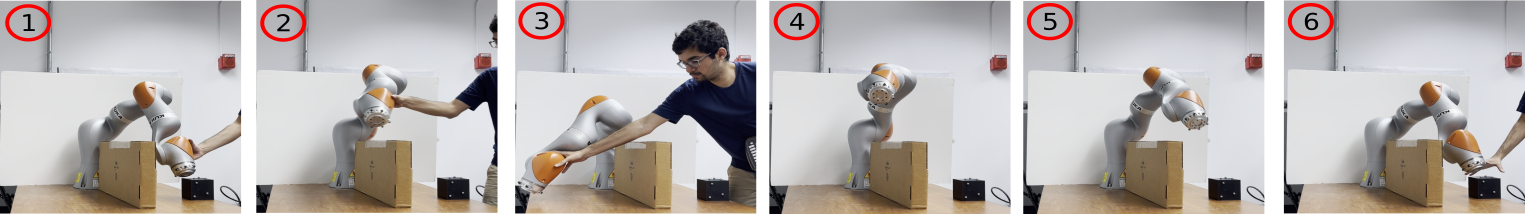}
\caption{KUKA recovering from pushes while avoiding an obstacle. The numbers on the images correspond to the sequence in the key frames.}
\label{fig:obstacle_push}
\vspace{-0.5cm}
\end{figure*}
where $f_{\text{fk}}(q_{t})$ is the forward kinematics function that returns the position of the end-effector (a highly nonlinear function of the states), $f_{\text{v}}(q_{t}, v_{t})$ computes the end-effector velocity using the end-effector Jacobian and $w_{1,2,\dots, 7}$ are cost weights that are tuned by the user. We use this cost in the bi-level optimization to train the QPNet as described in Sec.~\ref{sec:training}.  
This experiment is chosen to demonstrate that the predictions of the QPNet are satisfactory to be applied to a real robot when using sensor measurement that is normally used with trajectory optimization. For this experiment, the framework run on the robot is identical to the scheme presented in Fig.~\ref{fig:pipeline} with the only difference that the target reaching location is directly provided by \ac{MC}---there is no encoder network. 
% The inverse dynamics controller is run at a \SI{1}{\kilo \hertz} while the QPNet and \ac{QP} trajectory optimizer are updated at \SI{5}{\hertz} to generate new plans. If required, the update rate of the QPNet and \ac{QP} optimzer can be increased up to \SI{500}{\hertz} since the solve time for both is under \SI{2}{\milli \second}. 

The performance of the QPNet is shown in Fig.~\ref{fig:vicon_reaching} and the accompanying video. The images aligned with the shaded areas depicts the state of the robot at the respective times. As it can be seen, our method is able to generate motions to reach the cube accurately---the end-effector of the robot is in contact with the cube at key frames. The small error between desired and actual reaching position arises because the robot tries to reach the center of the cube which creates an offset. The generated trajectories are also smooth at run time, which is important for safe use on real robots. The QPNet is robust and stable to unforeseen pushes on the robot and brings back the robot to the location of the cube after the disturbance, even though the QPNet was never specifically trained for it. The robot is also very compliant while moving; this makes it safe to interact with, an advantage gained from constantly re-planning (\ac{MPC}).  

To systematically analyze the reaching performance of the QPNet, we run the reaching experiment in simulation for 1000 randomly generated points in the entire workspace. This includes a significant number of points outside the region in which training targets were created. The robot reached all the targets with a mean and standard deviation in reaching error of \SI{8}{cm} and \SI{3}{cm}, which is rather low given the size of the iiwa robot (\SI{1.5}{m}) and the small training set. Importantly, the method always generated stable trajectories that reached the target. The errors are rather small compared to other reported results for learning approaches \cite{bechtle2020curious}, even while considering targets outside of the training region.% as compared to other learning approaches \cite{bechtle2020curious}.}

\begin{figure*}[h!]
\vspace{0.2cm}
\centering
\includegraphics[scale=0.23]{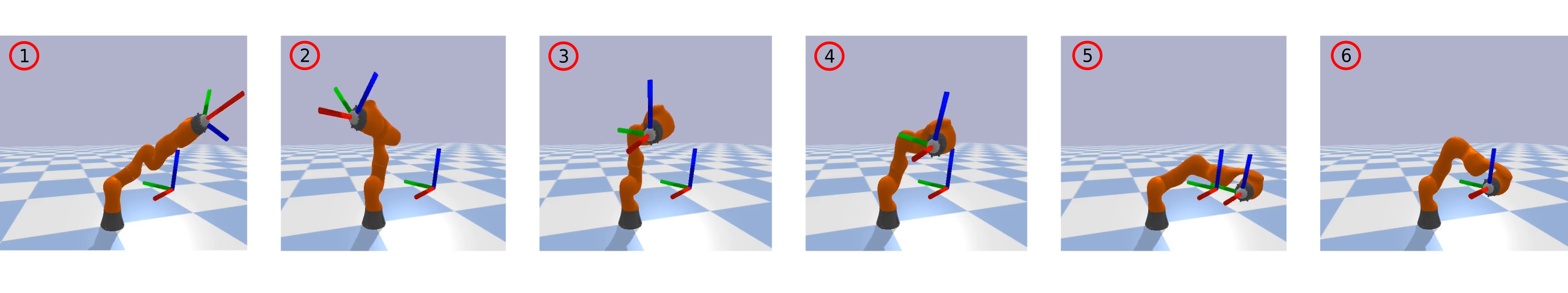}
\vspace{-0.3cm}
\caption{KUKA reaching a desired 6D pose in simulation. The numbers on the images correspond to the sequence.}
\label{fig:pose}
\vspace{-0.4cm}
\end{figure*}

\subsection{Reaching Task with Obstacle Avoidance}
We now increase the complexity of the task from just reaching a desired location in free space to also avoiding an obstacle while reaching. This is to demonstrate that the proposed method can achieve more complicated tasks while keeping the computation time unchanged. To achieve this, we now use a task loss for both obstacle avoidance and reaching. The obstacle avoiding task is encoded in the cost using a repulsive exponential potential
\begin{multline}
    \phi_{\text{obstacle}}(x, o) = \\ \sum_{t=1}^{N}w_{8}\exp\bigg[{-\Big(\big(f_{\text{fk}}(q_{t}) - o\big)^{T}A\big(f_{\text{fk}}(q_{t}) - o\big)-1\Big)/w_{9}}\bigg]\,,
\end{multline}
where $o$ is the location of the obstacle center, $w_{8}, w_{9}$ are tuning parameters that determine how strictly the obstacle should be avoided, and $A$ is a diagonal scaling matrix that determines the convex hull circumscribing the obstacle to be avoided. By changing the diagonal elements, the obstacle can take the shape of a sphere to a narrow capsule. The QPNet is now trained by solving the problem~(\ref{eq:task_loss}) using the combined task loss for avoiding and reaching $\phi_{\text{ar}} = \phi_{\text{reaching}}(x, g) + \phi_{\text{obstacle}}(x, o)$. In addition, we also add an intermediate goal to guide the end-effector around the obstacle as the potential field may result in the optimizer being stuck in local minima. This intermediate goal can be generated automatically by a RRT-based planner and is only needed to compute $G^{\star}, c^{\star}$. The trained QPNet is then used in the same pipeline as discussed in Sec.~\ref{sec:vicon}. Note that during execution, the via-point is not needed and trajectories avoiding the obstacle are automatically generated by the QPNet. %Consequently, the goal of the robot is now to avoid obstacles while reaching a desired target location.  
\begin{figure}
\vspace{0.2cm}
\centering
\includegraphics[scale=0.155]{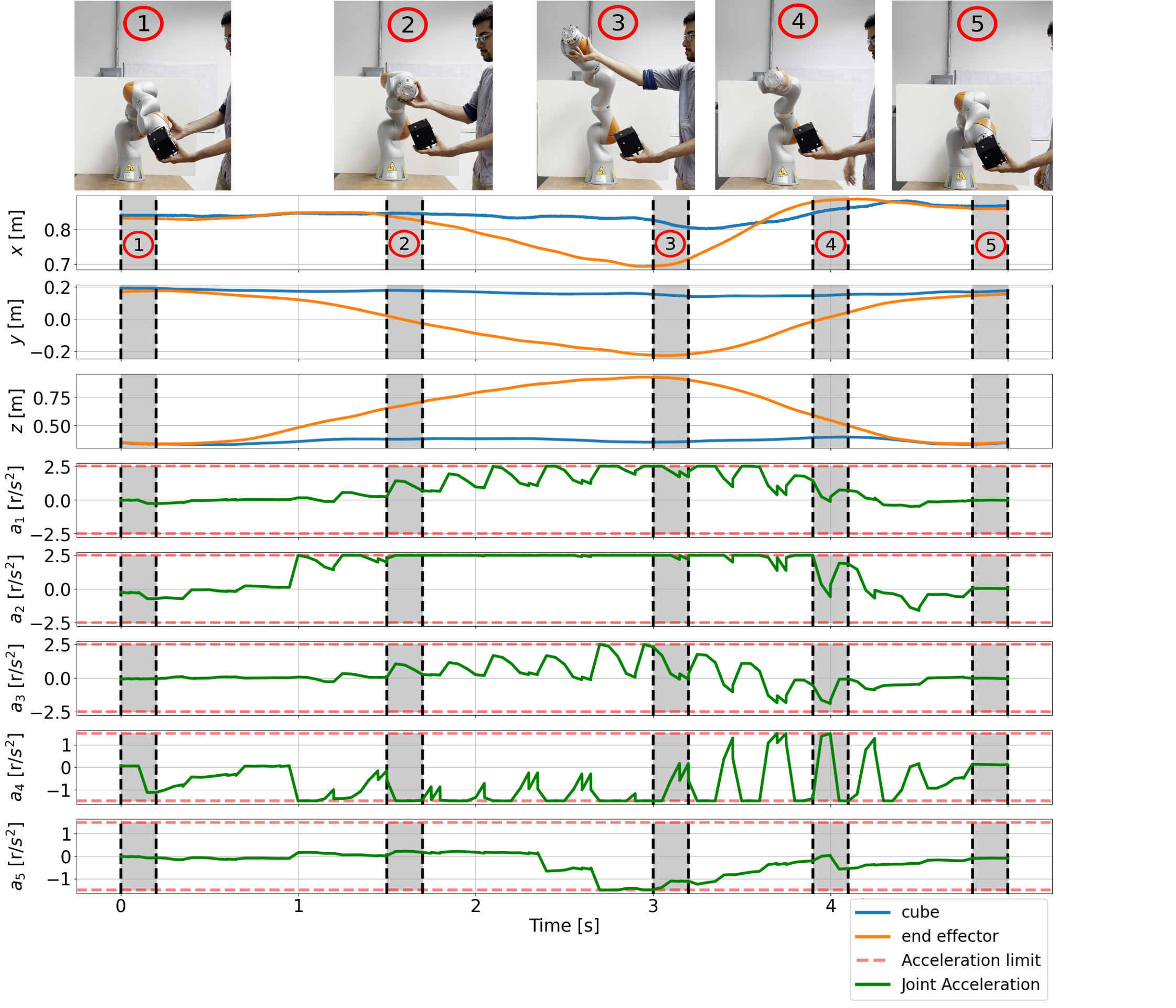}
\caption{KUKA recovering from pushes while using vision measurements in the MPC loop. The numbers on the images correspond to the state of the end-effector and cube in the shaded region.}
\label{fig:e2epush}
\vspace{-0.5cm}
\end{figure}
During execution, our approach successfully achieve this task without any increase in the computation time, since it still only requires one forward pass of the QPNet and solving one \ac{QP} at each \ac{MPC} time step. The robot reaches different cube locations that are changed online while avoiding the obstacle. The robot is robust to external perturbations, compliant and moves smoothly as presented in Sec.~\ref{sec:vicon}. A typical behaviour of the robot is shown in Fig.~\ref{fig:obstacle_push}: the robot is first pushed away from the desired target location (the black cube) and then released. The robot then moves back towards the cube while avoiding the obstacle (the brown box). More examples can also be found in the supplementary video. 

\subsection{6D Pose Reaching Task}
For tasks such as object grasping or insertion, it is desirable to control both the reaching position and orientation of the end effector. In this experiment, we show that by just adapting the nonlinear cost $\phi$ and retraining the QPNet we can control the 6D pose without any change to the computation time during deployment. The modified cost function is 
\begin{multline} \label{eq:posetask}   \phi_{\text{pose}}(x, g) = w_{6}\norm{\log_{6}(T(q) \ominus T_{\text{des}})} + w_{7}\norm{f_{\text{v}}(q_{N}, v_{n})} \\
            + \sum_{t=1}^{N-1} w_{1}\norm{\log_{6}(T(q) \ominus T_{\text{des}})} + w_{2}\norm{q_{t}} + w_{3}\norm{v_{t}} +
            w_{4}\norm{a_{t}} \\ + w_{5}\norm{a_{t} - a_{t-1}}\
\end{multline}
where $T(q)$ is the current 6D pose of the end effector, $T_{\text{des}}$ is the desired reaching pose and $\ominus$ is the difference operator in SE(3). We run this experiment in simulation, since it is difficult to show that the desired 6D pose is actually reached without a gripper on the robot. Figure \ref{fig:pose}, shows the KUKA robot reaching a desired 6D pose where the visual coordinate axis attached to the end-effector aligns with the desired one in space. More results are available in the attached video. 
% The same framework could be extended easily to pick \& place objects by including the gripper joints into the kinematics solver and adding a simple grasping cost to $\phi$.    

\subsection{Reaching Task with Vision}
We now present results when using a RGB-D camera in place of the \ac{MC} system as discussed in Sec.~\ref{sec:execution}. The encoder contains $9$ convolutional layers of width $64$, $64$, $128$, $128$, $256$, $256$, $256$, $512$, $512$ respectively, followed by $2$ fully connected layers with $512$ neurons each. All layers except the last one are activated with ReLUs and a max pooling layer is added after the $2^{nd}, 4^{th}, 5^{th}, 6^{th}$ and $7^{th}$ layer. The goal is again to reach the cube at different locations while using RGB-D images as sensor measurements. 
% With this architecture, the computation time of the image encoder, the QPNet, and the \ac{QP} totals to about $\SI{10}{\milli \second}$, which means that control frequencies up to \SI{100}{\hertz} could be achieved if needed.
%

Tracking performance is similar to that presented in Sec.~\ref{sec:vicon}. During execution, the robot remains very compliant yet precise thanks to online re-planning. The controller is able to reject unexpected pushes and remains stable. Figure~\ref{fig:e2epush} shows an example of the robot being pushed away from the target location (the cube) and the robot finds its way back to the cube when the external force is removed. The system remains stable during the push as shown by the smooth trajectory of the end-effector position and the desired joint acceleration obtained from the \ac{QP}. 
We further note that the joint accelerations remain bounded during the push, due to the constraints enforced by the \ac{QP} (the acceleration of joint 2 between \SI{1.5}{\second} to \SI{3}{\second} stays on the limit depicted by the horizontal dotted line). This demonstrates the advantage of using an optimizer that enforces hard constraints at all times, thus ensuring a certain level of safety. %which ensure safety of robots working in human environments.  

%===============================================================================

\section{Discussion}
\subsubsection{\ac{OCP} formulation} The common practice with \ac{MPC} techniques is to design a non-linear cost function that best describes the desired task and solve it online \cite{kleff2021high, pankert2020perceptive}. This is often achieved by using \ac{DDP} to solve the optimization problem. These solvers are custom implemented to reduce computation time. However, this limits the generality of the \ac{MPC} scheme since if the complexity of the task changes, the computation time may become too high for real-time planning. Here, we propose an alternate paradigm where we reformulate the \ac{OCP} as a bi-level optimization problem to generate and learn convex costs. This enables online computation regardless of the task complexity due to the fixed computation time of the neural network and \ac{QP}. In principle, our approach could then also be extended to other applications where the original \ac{OCP} is numerically challenging to solve such as in legged locomotion.  
\subsubsection{QPNet performance} In all our experiments (\ac{MC}, vision), the QPNet was able to reach any location in the workspace. It also generalized to targets that lied outside the region of the training data. The \ac{MPC} was compliant and safe even when pushed. The same framework could handle increased complexity in tasks  by simply modifying the high-level task loss, without any change to compute times. Our approach can be easily extended to more difficult tasks such as grasping or pick-and-place ny adapting the higher level loss accordingly. However, one downside is that precision of the reaching was of the order of a few centimeters, which is not suitable for high-precision tasks. Future work will include improving precision for manipulation tasks. 

% the method is currently suitable for low precision tasks such as reaching, pick-and-place large warehouse objects bigger than $\SI{8}{cm}$. Further work is needed to make it suitable for high precision activities like manipulation. 

% Currently, the lower-level \ac{QP} in our method only indirectly enforces torque limits by constraining joint accelerations. While this is a reasonable way to enforce limits~\cite{ponton2021efficient}, it may still violate these limits in some scenarios. Also, our method may still require the tuning of the high-level non-convex task loss $\phi_{\text{task}}$. While this is much easier than directly tuning the \ac{QP} cost, it can still be time-consuming at times. Exploring alternative ways of defining the task loss~\cite{bechtle2022model} might alleviate this difficulty. Finally, we only investigated reaching tasks and it remains to be shown whether force sensing can be added for tasks involving physical interactions and whether this leads to stable behavior during contact transitions. %This is something that will be explored in future work. 
%===============================================================================

\section{Conclusion} \label{sec:conclusion}
In this paper, we present an approach that can generate rich behaviours akin to non-convex trajectory optimization-based \ac{MPC} while only solving \acp{QP} online. This is achieved by training a neural network that constantly adapts the quadratic cost using sensor measurements. Due to this, we ensure real-time convergence and strict constraint satisfaction. Further, the use of a neural network to encode the cost function allows easy integration of high-dimensional sensory data into traditional trajectory optimization schemes while keeping the computation time invariant to the complexity of the task. We demonstrate the capability of this approach with diverse reaching tasks on a KUKA manipulator while fusing different sensor measurements.
%===============================================================================

\clearpage

%===============================================================================

\bibliography{ref}
\bibliographystyle{IEEEtran}
\end{document}